\documentclass[runningheads]{llncs}
\usepackage{graphicx}
\usepackage{comment}
\usepackage{amsmath,amssymb} 
\usepackage{color}

\titlerunning{Realizing Pixel-Level Semantic Learning in Complex Driving Scenes} 
\authorrunning{Xi Li, Huimin Ma, Sheng Yi, Yanxian Chen} 
\author{Xi Li, Huimin Ma, Sheng Yi, Yanxian Chen}
\institute{Tsinghua University, University of Science and Technology, Beijing}

\usepackage[width=122mm,left=12mm,paperwidth=146mm,height=193mm,top=12mm,paperheight=217mm]{geometry}

\usepackage{makecell}
\usepackage{multirow}

\begin{document}

\mainmatter
\def\ECCVSubNumber{1}  

\title{Realizing Pixel-Level Semantic Learning in Complex Driving Scenes based on Only One Annotated Pixel per Class} 

\maketitle

\begin{abstract}
Semantic segmentation tasks based on weakly supervised condition have been put forward to achieve a lightweight labeling process. For simple images that only include a few categories, researches based on image-level annotations have achieved acceptable performance. However, when facing complex scenes, since image contains a large amount of classes, it becomes difficult to learn visual appearance based on image tags. In this case, image-level annotations are not effective in providing information. Therefore, we set up a new task in which only one annotated pixel is provided for each category. Based on the more lightweight and informative condition, a three step process is built for pseudo labels generation, which progressively implement optimal feature representation for each category, image inference and context-location based refinement. In particular, since high-level semantics and low-level imaging feature have different discriminative ability for each class under driving scenes, we divide each category into "object" or "scene" and then provide different operations for the two types separately. Further, an alternate iterative structure is established to gradually improve segmentation performance, which combines CNN-based inter-image common semantic learning and imaging prior based intra-image modification process. Experiments on Cityscapes dataset demonstrate that the proposed method provides a feasible way to solve weakly supervised semantic segmentation task under complex driving scenes.
\end{abstract}

\section{Introduction}
\begin{figure}[!t] 
	\begin{center}
		\includegraphics[width=1\linewidth,trim = 0mm 0mm 0mm 0mm, clip]{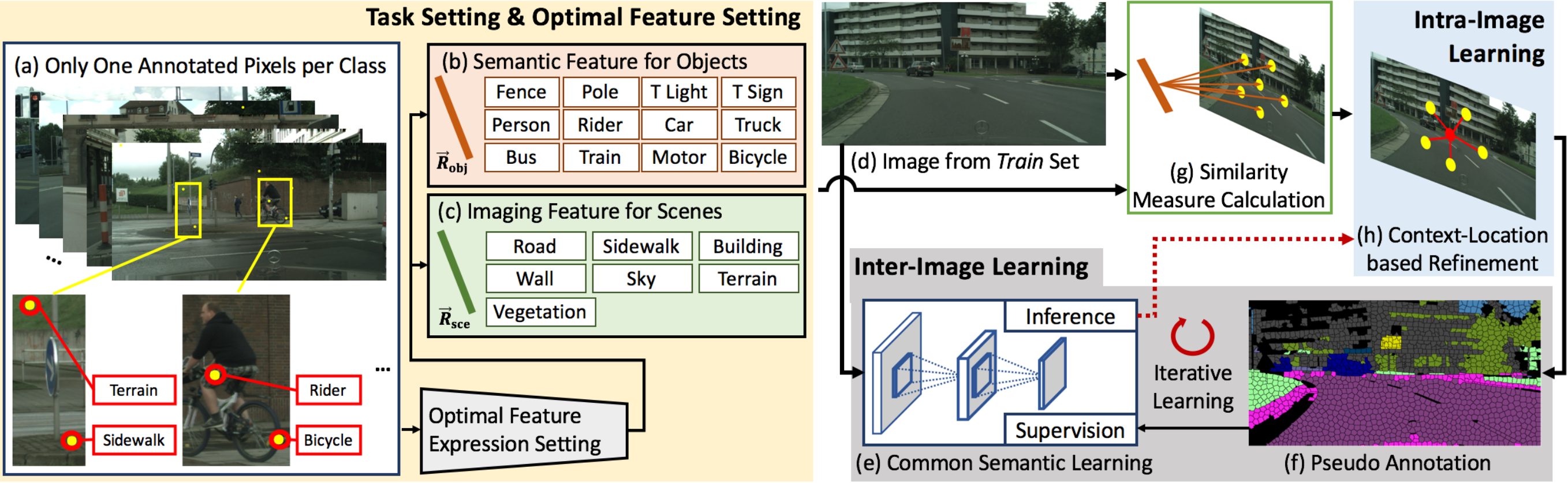}
		\vspace{-7mm}
		\caption{Illustration of the proposed weakly supervised semantic segmentation task and learning framework under complex driving scenes. (a) For a whole dataset, only one pixel which belongs to each class is labeled. To achieve pixel-level inference, we first divide each classes into "object" or "scene", and build optimal feature representation for both types: (b) semantic feature vectors $\vec{R}_{obj}$ for objects and (c) imaging feature vectors $\vec{R}_{sce}$ for scenes expression. Based on the representation, annotation can be inferred for each image (d) via similarity measure (g). Then, intra-image context relationship, as well as location prior are utilized for a further refinement process (h). Based on the modified labels (f), fully supervised based network can be utilized for common object and scene semantic learning (e). Moreover, an alternative iteration structure is presented to gradually improve the inference performance.}
		\label{fig:OverviewSim}
		\vspace{-9mm}
	\end{center}
\end{figure}

\begin{figure}[!t] 
	\begin{center}
		\includegraphics[width=1\linewidth,trim = 0mm 0mm 0mm 0mm, clip]{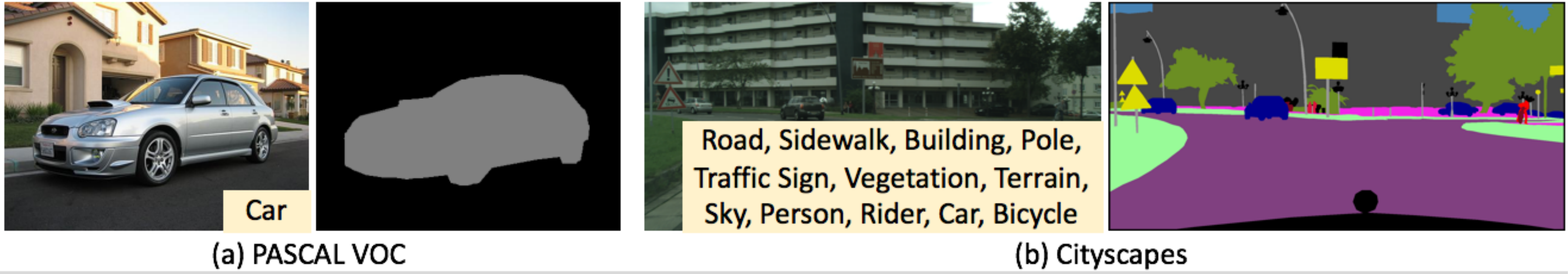}
		\vspace{-7mm}
		\caption{Example of images for semantic segmentation task in different datasets. In PASCAL VOC (a), images which contain single object categories take up a large percentage, while each type of background is not distinguished, so image-level labels could be used to obtain object regions which are responded via CAM \cite{zhou2016learning}. In contrast, image in Cityscapes (b) contains more complex scenes and a large amount of classes, which makes it hard to achieve high performance visual appearance based on image tags.}
		\label{fig:exp}
		\vspace{-9mm}
	\end{center}
\end{figure}

Weakly supervised learning is a task that provides insufficient annotations in response to inference requirements. For semantic segmentation under weakly supervised condition, the task is set to achieve pixel-level semantic learning while only bounding box- \cite{dai2015boxsup}, \cite{khoreva2017simple}, \cite{song2019box}, scribble- \cite{lin2016scribblesup} or image- level \cite{ahn2018learning}, \cite{huang2018weakly}, \cite{wei2017object}, \cite{shimoda2016distinct}, \cite{wei2017stc}, \cite{lee2019ficklenet} labels are provided in training set. Meanwhile, most existing works are based on images under simple scenes. Taking the PASCAL VOC \cite{everingham2010pascal} dataset as an example, most of training samples only include single or a few object categories, meanwhile objects usually occupy a significant proportion of region in images. Under the condition, it is possible to learn objects' visual representation with image-level tags via classification network \cite{zhou2016learning}, and further realize pseudo labeling and learning. However, when it comes to complex driving scenes, such as Cityscapes \cite{cordts2016cityscapes} dataset, since each image contains a large number of classes, meanwhile several categories appear almost in every samples (such as road, sky, etc.), image tags are difficult to be utilized for class activation directly, labeling process is no longer sufficiently lightweight as well. Figure~\ref{fig:exp} illustrates the comparison of images and annotations in different datasets.

To provide lightweight and reasonable initial labeling setting, in this paper, a new semantic segmentation task is put forward for complex driving scenes, in which only one annotated pixel for each class is provided. It means that only $C$ pixels are labeled for the whole dataset containing $C$ categories. Figure~\ref{fig:OverviewSim} (a) shows the annotation form under the task. Note that the selection of initial labeled pixels is done manually, and the annotated pixels should be highly discriminative to humans, rather than completely random.  We argue that this labeling way is necessary for stable initial feature encoding, and is also lightweight in practical applications. Compared to image tags based setting, the proposed task removes redundant information that is difficult to utilize, while provides the simplest but comprehensive initial label for each category. 

Under the proposed task setting, a two-stage learning framework is established, which first generates pseudo labels for each image and then implements CNN-based learning process. When facing driving scenes, it can be found that each class has variable expression capabilities with different feature layers. Specifically, semantic feature encoding via CAM \cite{zhou2016learning} model pre-trained on ImageNet could provide stable expression for objects, while each type of "scene" could be represented as limited types of color and texture combinations. Meanwhile, intra-image context and location prior are also discriminative for driving scenes.

Based on the observation, a three-step process is presented for pseudo labeling. First, in order to provide intra-class stable and inter-class discriminative expressions, optimal feature representation setting is proposed. Here, based on the only one annotated pixel, each category is divided into object or scene and represented via semantic or imaging feature pools respectively. Then, by establishing similarity measure between each pair of pixels, inference process can be operated for each image. Moreover, context and location prior in driving scenes are introduced for intra-image relationship modeling and annotation refinement. Based on the generated pseudo labels, alternative iteration learning structure is established, which gradually improves the performance of common semantic learning. Figure~\ref{fig:OverviewSim} shows the proposed learning framework. With only one annotated pixels per class, the proposed method could finally achieve pixel-level semantic learning and inference.

The main contributions of this paper are three-fold.

First, a new task for weakly supervised semantic segmentation under complex driving scenes is put forward, in which only one annotated pixel per class is provided. The initial labeling process follows a more lightweight, reasonable and applied setting.

Second, a novel pseudo labeling strategy and alternative iteration learning structure is established. High quality pseudo annotations could be obtained via making full use of semantic and imaging features for objects and scenes representation. Further, by alternately iterating context-location prior based on intra-image refinement and CNN-based inter-image common class semantic learning, segmentation results can be obtained and gradually optimized.

Third, a complete framework to solve weakly supervised tasks under complex driving scenes is put forward. Experimental results on Cityscapes demonstrate both rationality of the task setting and effectiveness of the proposed method.

The rest of paper is organized as follow: Sec.~\ref{sec: relate} illustrates related works. In Sec.~\ref{sec: overview}, basic framework of the proposed method is introduced, followed by the details of pseudo label generation and alternative iteration learning structure in Sec.~\ref{sec: pseudo} and ~\ref{sec: CNN} respectively. We detail the experiments and analysis in Sec.~\ref{sec: exp}, while conclusion is presented in Sec.~\ref{sec: cons}.

\section{Related Work}\label{sec: relate}
\subsection{Fully-Supervised Semantic Segmentation}
Semantic segmentation task under fully supervision provides pixel-level annotations for all training samples. Under the circumstance, a large amount of CNN based methods have been presented followed by FCN \cite{long2015fully}, which establish end-to-end learning frameworks and achieve decent performance. DeepLab \cite{chen2018deeplab} together with its variants, introduce dilated convolutions \cite{chen2017rethinking}, \cite{wang2018understanding}, conditional random fields (CRFs) \cite{krahenbuhl2011efficient}, \cite{zheng2015conditional} and other units which operate optimization for segmentation results. Although method based on fully annotations cannot be applied to weakly supervision based tasks directly, the framework can still support to mine common object feature \cite{wang2018weakly}, \cite{li2019weaklier}. Thus, based on incomplete pseudo labeling information, fully supervision based segmentation network structure is utilized in the proposed method for inter-image common class semantic learning.

\subsection{Weakly-Supervised Semantic Segmentation}
Under weakly supervised conditions, image-level based task gains more attention. For images with simple scenes, multi-instance learning \cite{pinheiro2015image}, \cite{saleh2016built} based on a classification network could provide coarse semantic segmentation results via image tags. On the basis of it, methods with localization based framework \cite{wang2018weakly}, \cite{ahn2018learning}, \cite{huang2018weakly}, \cite{kolesnikov2016seed}, \cite{papandreou2015weakly}, \cite{wei2017object}, \cite{wei2017stc}, \cite{lee2019ficklenet} achieve results with higher performance and more detailed regions. Specifically, these methods first generate object seeds as pseudo supervised annotations, and further imply pixel level learning and inference based on fully supervised structure. 

In order to achieve better performance, localization based methods focus on pseudo label generation and learning process respectively. AffinityNet \cite{ahn2018learning} defines an affinity metric which describes relationship between intra-image pixel pairs, while heat map fusion strategy is put forward in \cite{ge2018multi} to combine features from different layers. Besides, three losses are set in SEC \cite{kolesnikov2016seed}, which provide guidance for pseudo supervised seeds expansion and constrain. Similarly, DSRG \cite{huang2018weakly} is a method to implement object region growing iteratively based on deep features. Meanwhile, these methods are all based on initial object seeds which are generated based on CAM \cite{zhou2016learning} and its optimized versions. In complex scenes, since image tags based learning methods can not directly provide accurate visual appearance for each category, image tag based task setting becomes inappropriate, while the methods are difficult to achieves satisfactory performance. Therefore, we establish a new task for complex driving scenes, which is only providing one annotated pixel for each class. By leveraging semantics, imaging, context feature and location prior, a framework with the two-stage localization based learning structure is established and effectively work under the proposed task.

Besides, a few studies also pay attention to weakly supervised learning under complex scenes. Li \textit{et al.} \cite{li2018weakly} present task setting which provides box-level annotations for things, while stuff is labeled with image tags, a learning framework process the two type of category separately. Saleh \textit{et al.} \cite{saleh2017bringing} propose a solution to multi label background learning based on video sequences. In addition, to achieve semantic segmentation on new datasets, \cite{zou2018unsupervised}, \cite{zou2019confidence} focus on establishing common relationship between source and target domain. In contrast, annotations in our proposed task only rely on still images with only one labeled pixel per class from a single dataset, which is more lightweight and informative. Based on limited annotations, in order to provide discriminative expression ability for each category, we pay attention to describe optimal feature representations. Inspired by \cite{li2018weakly}, we also assign each class to two types: "object" or "scene", and build different representations, inference and refinement processes respectively. Based on the strategies, satisfactory pseudo labeling and semantic segmentation performance can be implemented.

\section{Overview of the Proposed Method}\label{sec: overview}
For the proposed weakly supervision task under driving scenes, a two-stage framework is established. Based on only one annotated pixel per class, the first process is to achieve feature representation for each class, followed by the process of pseudo label inference and optimization for images. Then, segmentation network is introduced in an alternative iteration process to perform common semantic learning with a number of annotated samples.

To achieve initial pseudo annotations for training samples, three steps are undergone. First, optimal feature representation is set to express each category based on the one annotated pixel. Here, CAM \cite{zhou2016learning} model pre-trained on ImageNet is introduced to extract high-level semantic features, while color and texture are encoded for imaging feature expression. Under driving scenes, it can be observed that for a type of categories (such as car, person), semantic feature from a same class tends to have small intra-class variances, while imaging features for individual samples are diverse and hard to be built as a unified expression. While for the remains (\textit{i.e.} sky, vegetation), imaging feature of regions with same label can be described as limited number of concentrated distributions, but stable semantic response cannot be provided via CAM model trained on ImageNet. Based on the observation, each category is divided into one of the types according to intrinsic attributes, which are named as object and scene. With the division, feature vectors based on semantic and imaging expression are established for objects and scenes representation respectively. Note that when setting the optimal feature expression for each class, not only features of annotated pixel are observed, but also intra-image context relationship is considered to obtain an integrated expression. 

Utilizing the optimal feature based representations, the similarity measure of each pixels pair can be calculated on each category, including both pixel pairs from a same image or different samples. Initial pseudo annotations of each image can be obtained via similarity calculation. Further, context-location refinement (CLR) strategy is presented for optimizing the results via intra-image prior analysis. Specifically, global-local saliency and edge are encoded to measure the context relationship for confident scores modification of each region. Meanwhile, location prior under driving scenes are defined to restrict regions where each category could exist. As pixels with unreasonable annotations can be modified, pseudo labels with higher precision and recall can be obtained.

In the second stage, an alternative iteration structure is used to achieve semantic segmentation. Based on the samples with pseudo annotations, CNN based segmentation network is used to achieve common object and scene semantic learning. Furthermore, each sample could be re-inferred via a trained model, then CLR unit is introduced for generating and optimizing pseudo labels for a new round. Through alternative iteration, inter-image common semantic learning process and intra-image refinement can be fully integrated, which enables the performance to be gradually improved. The complete framework is shown in Figure~\ref{fig:OverviewSim}. Note that only model trained in the last iteration is used for inference, so it can be efficient for practical application.

\section{Pseudo Label Generation} \label{sec: pseudo}

\begin{figure*}[!t] 
	\begin{center}
		\includegraphics[width=1\linewidth,trim = 0mm 0mm 0mm 0mm, clip]{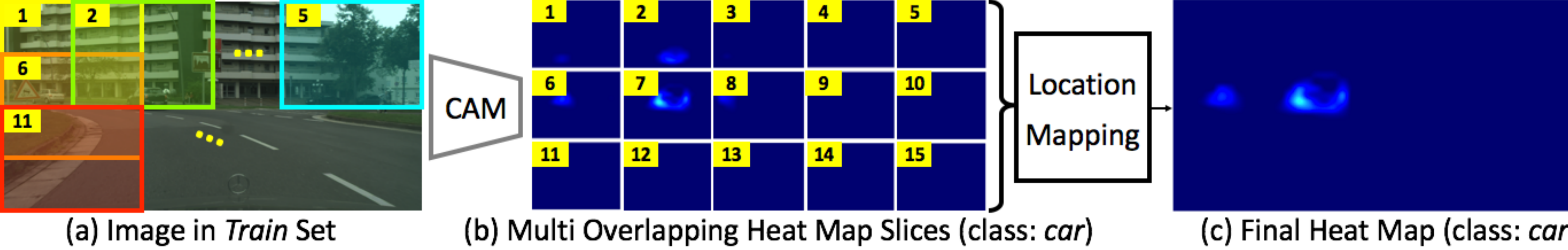}
		\vspace{-7mm}
		\caption{The framework of multi overlapping slices fusion (MOSF). Give an image (a), 15 overlapping slices are cropped for heat map slices encoding (b). The final heat map (c) is generated based on location mapping relationship. While in actual operation, final heat maps are recorded as intermediate results. The visual result shown here has undergone additional similarity calculation and resize operation.}
		\label{fig:MSC}
	\end{center}
	\vspace{-9mm}
\end{figure*}

\subsection{Semantic and Imaging Feature Encoding}

To achieve representation for each class and similarity measure of different regions and images, several methods are utilized to encode low-level imaging and high-level semantic features for images. Specifically, color, texture, edge, and superpixels are generated referred to \cite{li2018saliency}, while saliency maps are calculated by DRFI \cite{jiang2013salient}. Considering that regions belonging to a same instance should have similar saliency value under arbitrary observations, so not only saliency maps by encoding integrate image are generated, but also fragments based on two cropping types are inferred and recorded. Based on the strategy, observations with global and local views can be both introduced for saliency similarity measuring. Note that the cropping types is to ensure that relationships of adjacent superpixel pairs can be fully considered, so any partitioning form that meets the requirement can be used as substitution. Figure~\ref{fig:ilm} (b)-(e) shows an example.

\begin{figure}[!t] 
	\begin{center}
		\includegraphics[width=1\linewidth,trim = 0mm 0mm 0mm 0mm, clip]{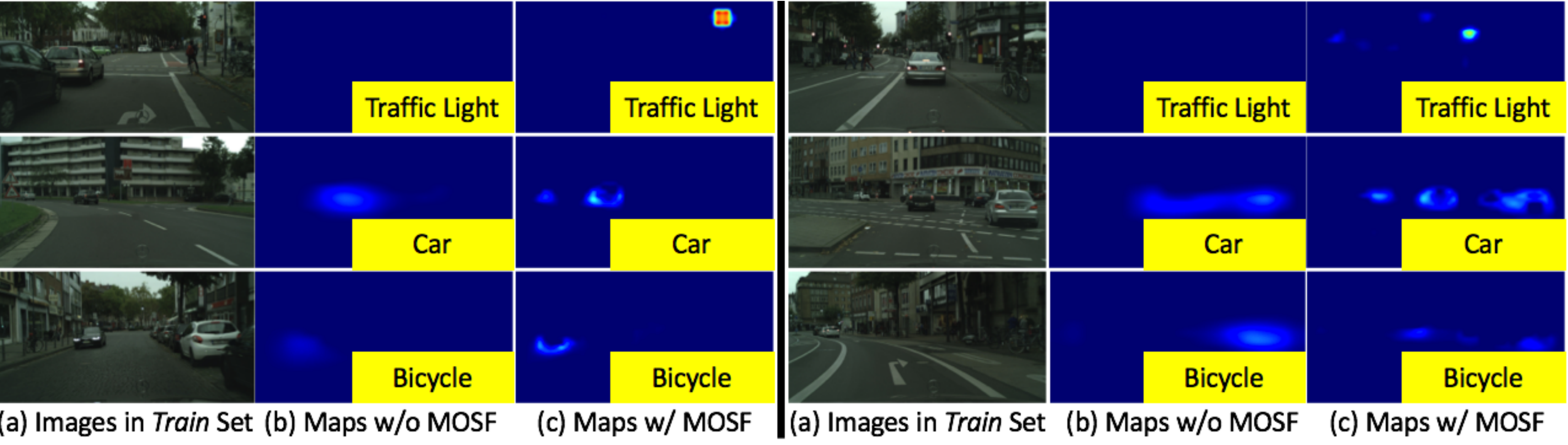}
		\vspace{-7mm}
		\caption{Examples of heat maps via MOSF. Images (a) contains different objects (T light, car, bicycle) are displayed. Compare to heat maps (b) generated by encoding a whole image, results provided by MOSF (c) achieve more detailed object regions.}
		\label{fig:vis_msf}
	\end{center}
	\vspace{-9mm}
\end{figure}

Besides, since the provided annotations are not sufficient to support fine-tuned operation on target dataset, CAM model \cite{zhou2016learning} pre-trained on ImageNet is used to extract high-level semantic feature. In this process, we do not encode for entire image directly, but divide it into several overlapping slices and generate heat map for each piece separately. The final semantic maps are generated via multi overlapping slice fusion (MOSF). Figure~\ref{fig:MSC} illustrates the basic process of MOSF. In particular, for an image with a size of $l \times w$, we divide it into $15$ overlapping $l/2 \times w/3$ slices, whose coordinates of top left corners are $(l\cdot{m/4}, w\cdot{n/6})$ ($m=0, 1, 2$, $n = 0, \cdots, 4$). For each slice, heat map is encoded via CAM \cite{zhou2016learning}. Further, Euclidean distance between pixels and slices' centers is calculated, then semantic feature of each pixel is represented by response vector from the same position in its nearest slice. Figure~\ref{fig:vis_msf} shows several visual examples. It can be admitted that, for images under complex driving scenes with a large amount of categories, MOSF could not only provide more accurate and comprehensive details, but also reduce the influence of objects' truncation via slices overlapping.

With the generated feature maps, superpixels are used as region encoding and representation units. For a superpixel $sp_i$, the semantic feature is defined as $\vec{H}(sp_i)$, while $\vec{C}(sp_i)$ and $\vec{T}(sp_i)$ represent for color and texture respectively. Saliency under global and local views are recorded as $\vec{Sal_g}(sp_i)$ and $\vec{Sal_{l_*}}(sp_i)$. Here, $\vec{H}(sp_i)$ is an 1000-$d$ vector obtained by averaging heat map response of pixels included in $sp_i$. Meanwhile, 32-$d$ vector is established as expression on other feature maps that are normalized, which is statistical distribution in interval $[0,1]$ of the contained pixels' value. Note that three channels are used for color description, so $\vec{C}(sp_i)$ is with 96-$d$. $E(i,j)$ is defined to describe the similarity measure of $sp_i$ and $sp_j$ on edge map following by \cite{li2019weaklier}, whose value is in the range of $[0,1]$, while larger value corresponds to higher similarity measure.

\subsection{Optimal Feature Setting for Object and Scene}

\begin{figure*}[!t] 
	\begin{center}
		\includegraphics[width=1\linewidth,trim = 0mm 0mm 0mm 0mm, clip]{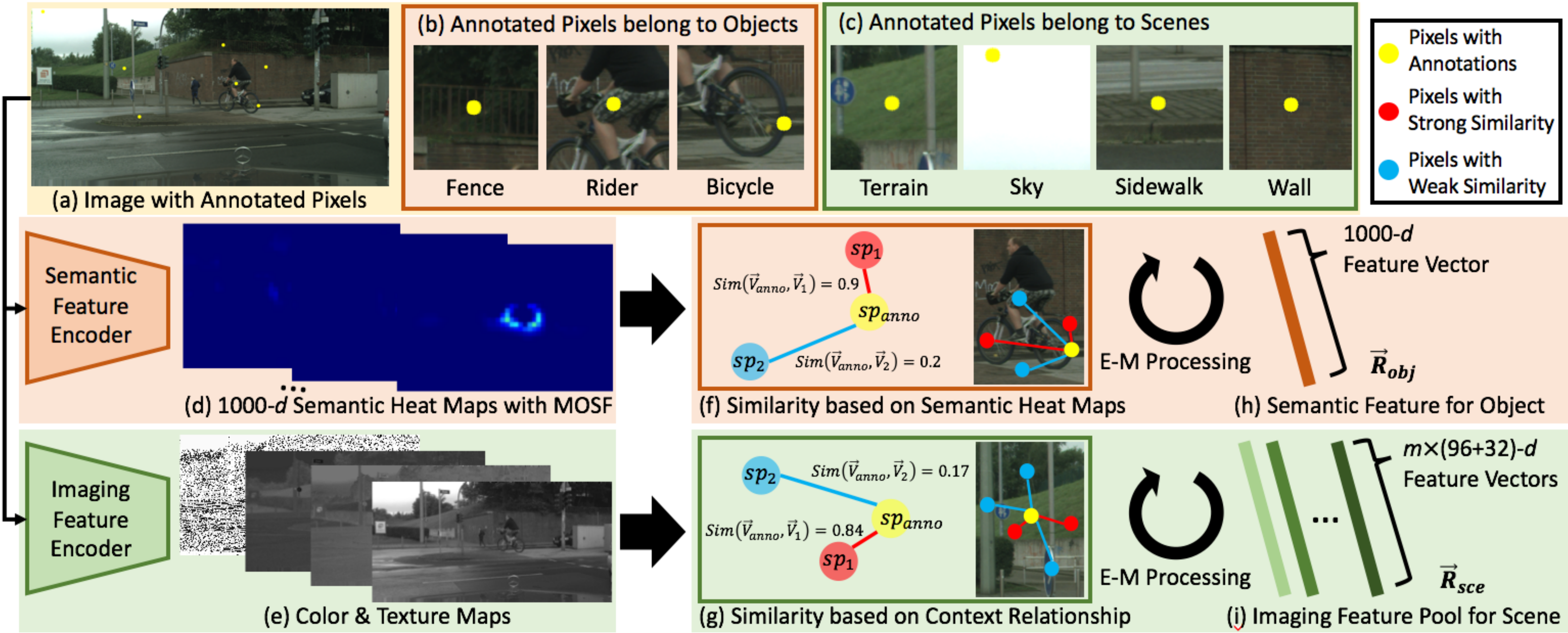}
		\vspace{-7mm}
		\caption{Overview of optimal feature setting for each class. For an image (a) contains annotated pixels which are divided to object (b) or scene (c), high level semantic heat maps base on MOSF (d), together with imaging feature maps (e) are encoded. For class which defined as object, similarity (f) between each superpixel based on semantic feature is calculated, while global-local saliency and edge features are used for the measure of scenes (g). Based on the feature vector of labeled pixel and similarity measure with other regions in image, E-M process is utilized to generate semantic and imaging feature based object (h) and scene (i) representation ($\vec{R}_{obj}$, $\vec{R}_{sce}$) respectively.}
		\label{fig:OFS}
	\end{center}
	\vspace{-9mm}
\end{figure*}

In order to infer the category information of each region and achieve pseudo labeling, each class should be expressed as feature vectors which could provide stable description based on the one annotated pixel. Since different classes have divergent discriminative ability on semantic and imaging features under driving scenes, we divide all classes into object and scene based on intrinsic properties. Then semantic features and imaging features are utilized to create optimal feature representation for the two types respectively.

Separately, for object feature expression, heat maps encoded by MOSF are introduced. For image $I$ contains annotated pixel with object category $C_{obj_*}$, semantic feature vectors $\vec{H}(sp_*)$ of superpixels included in $I$ are provided. While, $\vec{H}(anno)$ is defined as the response vector of superpixel which includes the labeled pixel. Note that each response vector is normalized. Further, E-M strategy is used to generate the feature representation. $\vec{H}(anno)$ is used as initial center $\vec{M}_g$. The similarity measure of $\vec{M}_g$ and semantic feature vectors of other superpixels are calculated via eq (1).

\vspace{-2mm}
\begin{equation}
Sim(\vec{X}_i, \vec{X}_j) = \sum_{n=1}^{N} min(x_{in}, x_{jn})
\end{equation}

Here, larger result indicates a higher similarity measure of $\vec{X}_i$ and $\vec{X}_j$, while $\vec{X_i}$ and $\vec{X_j}$ are represented by $\vec{M}_g$ and $\vec{H}(sp_*)$ respectively in this specific operation. Then, top $1\%$ vectors with highest similarity scores are recorded in feature group $\Omega_g$, while optimization function (2) is designed to perform the re-inference process of $\vec{M}_g$.

\vspace{-4mm}
\begin{equation}
\begin{split}
\max \sum_{g=1}^{G} \sum_{\vec{X}_i \in \Omega_g}\sum_{n=1}^{N} \min (x_{in}, m_{gn}) \\  s.t. \ \ ||\vec{M}_g||_1 = 1, \ \ g = 1,\cdots G \ \ \ \
\end{split}
\end{equation}

Particularly, single feature vector is used for each object's representation. Therefore, the total number of clusters $G$ is set to $1$ under this condition. By updating $\vec{M}_g$ and vectors in $\Omega_g$ alternately, the final generated $\vec{M}_g$ is used as feature representation $\vec{R}_{obj_*}$ for object category $C_{obj_*}$.

Meanwhile, the similar process is operated to obtain scene's representation. For an image contains annotated pixel with scene category $C_{sce_*}$, color, texture, global-local saliency and edge feature vector for each superpixel $sp_i$ need to be generated. As a same scene could have several kinds of color and texture representations that differ from each other, saliency and edge are utilized for context similarity measure $Sim_{c}$ calculation, in order to discover the neighborhoods of superpixel $sp_{anno}$ which contains the pixel labeled by $C_{sce_*}$, as eq (3).

\vspace{-2mm}
\begin{equation}
Sim_{c}(i,j) = E(i,j)\cdot{Sim_g(i,j)}\cdot{\max\limits_{k = 1,2}(Sim_{l_k}(i,j))}
\end{equation}
Where $Sim_{c}$ is a matrix whose element $Sim_{c}(i,j)$ is the context similarity measure of $sp_i$ and $sp_j$. $Sim_g$ and $Sim_{l_*}$ are calculated by eq (1) via global and two local saliency features respectively. 

To obtain the neighborhoods of $sp_{anno}$, superpixels which have context similarity values greater than $0.5$ with $sp_{anno}$ are selected. In addition, similarity values based on color $\vec{C}(sp_*)$ and texture $\vec{T}(sp_*)$ between these superpixels are calculated via eq (1) respectively, while the product of these two measure is used to represent imaging feature similarity for each pairs. Superpixels whose values greater than $0.5$ are divided into $G$ groups, then the average of vectors from each group $\Omega_g$ is set as initial center $m_g$. Function (2) is introduced for E-M process. Finally, $G$ vectors are recorded as imaging feature pool $\{\vec{R}_{sce_{*}}\}$ to represent $C_{sce_*}$, where each 128-$d$ vector $\vec{R}_{sce_*}$ is a concatenation of color and texture feature. Note that objects can be well described via single semantic feature vector, while since a same scene may contain multiple imaging representations, limited number of feature groups are essential to achieve comprehensive description. Figure~\ref{fig:OFS} illustrates the framework of optimal feature setting.

\subsection{Image Inference via Similarity Measure}

Based on the feature representations, similar measure on each category can be inferred for each pixel in images. Here, image is divided into superpixels $sp_i$, while the semantic and imaging features ($\vec{H}(sp_i)$, $\vec{C}(sp_i)$ and $\vec{T}(sp_i)$) are encoded as well. In this process, objects and scenes are still considered separately. When inferring the relationship of $sp_i$ to object class $C_{obj_*}$, eq (1) is used based on semantic feature. Here, $\vec{X}$ is instead by $\vec{R}_{obj_*}$ and $\vec{H}(sp_i)$. Similarly, for scene which is expressed by imaging feature pool with $G$ groups of $\vec{R}_{sce_{*}}$, similarity measure between color-texture of $sp_i$ and each $\vec{R}_{sce_{*}}$ is calculated, while the maximum value is record as score. Then, $\vec{SM}_{obj}(sp_i)$ and $\vec{SM}_{sce}(sp_i)$ are established to record $sp_i$'s similarity measure on each class, whose dimensions are the number of object and scene types respectively.

Simply, category with highest similarity measure could be set as label for each superpixel. While, in addition to inference process from category representation to images via feature similarity measure, intra-image prior can also guides label generation. Therefore, $\vec{SM}_{obj}$ and $\vec{SM}_{sce}$ are recorded as intermediate results, which are waiting for post processing based on intra-image information.

\subsection{Context-Location based Refinement}

\begin{figure}[!t] 
	\begin{center}
		\includegraphics[width=1\linewidth,trim = 0mm 0mm 0mm 0mm, clip]{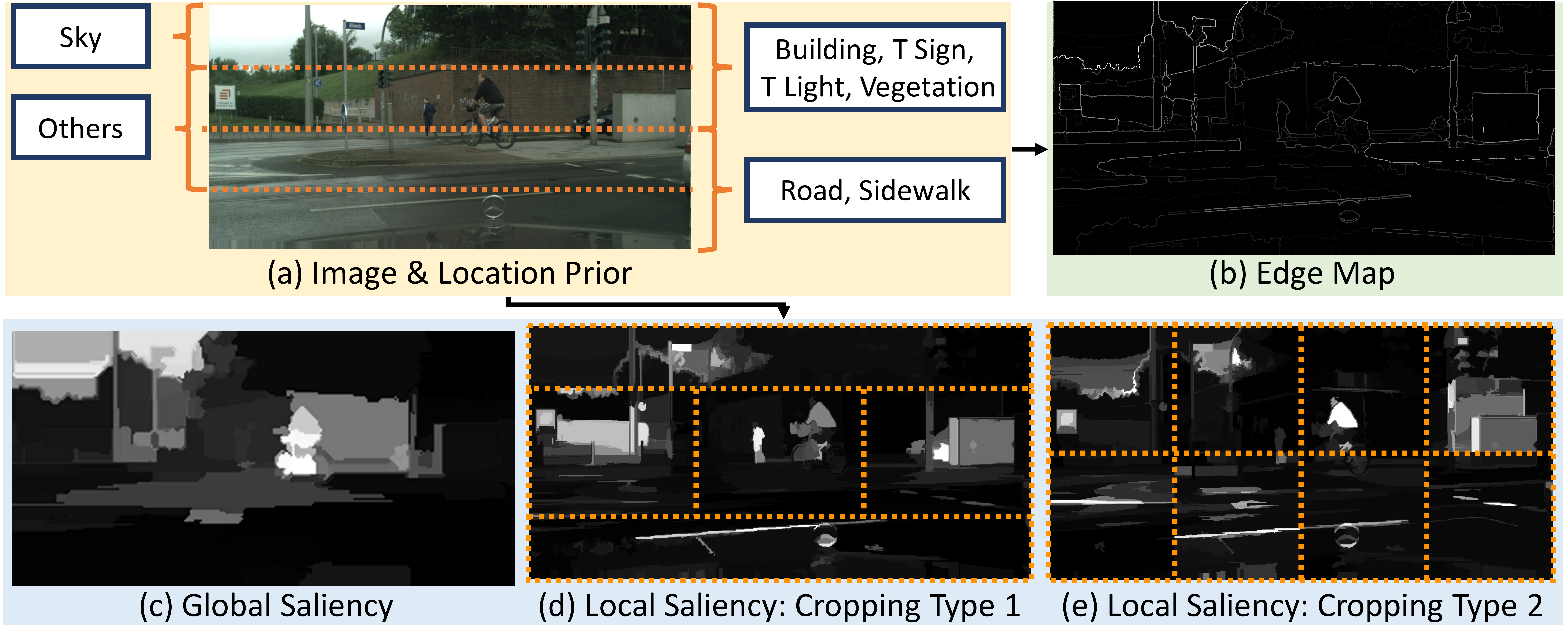}
		\vspace{-7mm}
		\caption{Settings of context-location based refinement (CLR). Each class is specified to exist in restrict region (a). Edge map (b), as well as global-local saliency maps based on different cropping rules (c)-(e) are defined for measuring the context similarity.}
		\label{fig:ilm}
	\end{center}
	\vspace{-9mm}
\end{figure}

To introduce the intra-image relationship of superpixel pairs and location prior under driving scenes, context-location based refinement (CLR) unit is designed to implement pseudo label generation. For an image contains $K$ superpixels, $K$ groups of $\vec{SM}_{obj}$ and $\vec{SM}_{sce}$ vectors can be obtained, which can be expressed as $M_{obj}$ and $M_{sce}$, where $M_{*}=[\vec{SM}_{*}(sp_1), \cdots, \vec{SM}_{*}(sp_K)]$. In addition, based on global-local saliency and edge feature, context similarity measure of each superpixel pair can be described as matrix form via eq (3). Then, the saliency and edge based refinement process for all vectors $\vec{SM}_{*_{r}}$ can be describes as eq (4) based on weighted calculation.

\vspace{-2mm}
\begin{equation}
M_{*_{r}}=[\cdots, \vec{SM}_{*_{r}}(sp_i), \cdots]=M_*\cdot{Sim_{c}}
\end{equation}

Here, each column of $Sim_{c}$ is normalized before operation, while refinement for objects and scenes are performed separately. Through the calculation, context of whole image is introduced for each superpixel's inference process, while superpixel pairs with higher value have greater influence on each other.

Meanwhile, as the spatial distribution of each class is restricted under driving scene, location prior is set as shown in Figure~\ref{fig:ilm} (a). Given an image, we divide it into four equal regions from top to down. For each region, we restrict the range of categories which participate in inference process. Specifically, for superpixel $sp_i$ locates in a certain region, only scores of categories within the restriction are exacted from similarity measure vectors $\vec{SM}_{obj_{r}}(sp_i)$ and $\vec{SM}_{sce_{r}}(sp_i)$. The highest score $Sco_{obj}(sp_i)$ and its corresponding object class $Cls_{obj}(sp_i)$ are recorded as pseudo tags for object attribute inference. Similarly, $Sco_{sce}(sp_i)$ and $Cls_{sce}(sp_i)$ are record for scene. In addition, $Cls_{*}$ with low scores are labeled as 255 (the corresponding region is ignored when training). In experiments, thresholds for both object and scene are set to $0.05$. Then, considering that objects tend to occupy small proportion of pixels in driving scenes, in order to ensure that each category obtains enough training samples, final pseudo label for each superpixel $Cls(sp_i)$ is defined as eq (5). Pixel-level annotations are generated through location mapping for images.

\begin{figure}[!t] 
	\begin{center}
		\includegraphics[width=1\linewidth,trim = 0mm 0mm 0mm 0mm, clip]{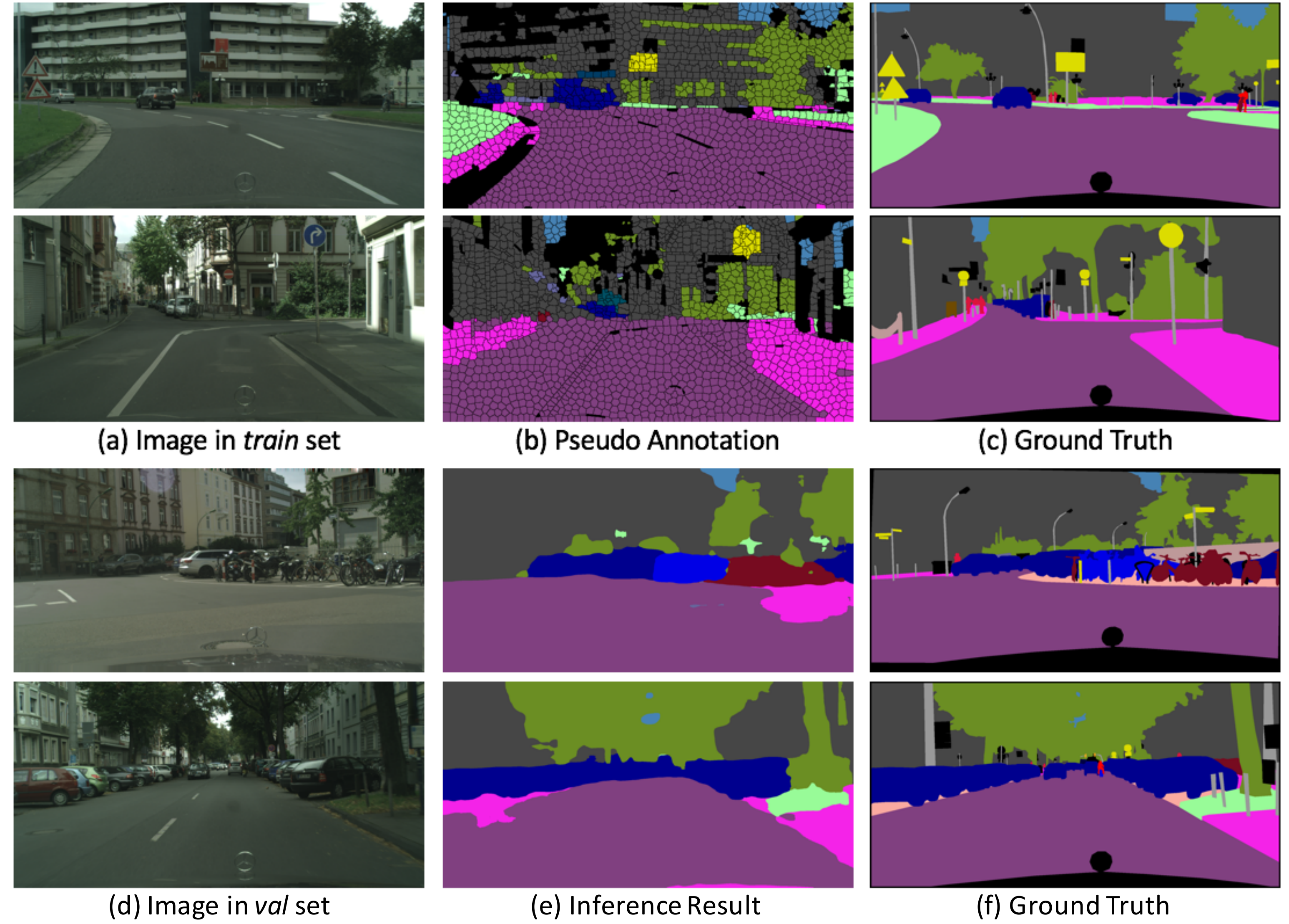}
		\vspace{-7mm}
		\caption{Qualitative results on Cityscapes \textit{train} and \textit{val} set.}
		\label{fig:vis_result}
	\end{center}
	\vspace{-9mm}
\end{figure}

\vspace{-2mm}
\begin{equation}
Cls(sp_i)=\left\{
\begin{aligned}
Cls_{obj}(sp_i) &\ \ \  Cls_{obj}(sp_i)\not=255 \\\
Cls_{sce}(sp_i) &\ \ \  else
\end{aligned}
\right.
\end{equation}

Through CLR strategy, as intra-image context information as well as location prior achieves fully utilization, more accuracy labeling results are provided. 

\section{Alternative Iteration based Semantic Segmentation} \label{sec: CNN}

Based on a large amount of acquired samples with pseudo annotations, CNN based fully supervised learning framework is applied for common object and scene semantic learning. In experiments, deepLab-v2 \cite{chen2018deeplab} based on VGG-16 is introduced as backbone. In practice, the framework can be replaced by variable semantic segmentation models. Then, for image $I$ from \textit{train} set, re-inferred annotation $Cls(x, y)$ for each pixel $(x,y)$ is provided by a trained model. Similarity measure vector $\vec{SM}(sp_i)$ of each superpixel $sp_i$ from $I$ are recorded as the class distribution of all contained pixels, as eq (6).

\vspace{-2mm}
\begin{equation}
sm_j= \frac{\sum_{(x,y)\in sp_i} Cls(x,y) = j}{\sum_{(x,y)\in sp_i} 1}, \ \  j = 1, \cdots, C
\end{equation}
Where $sm_j$ is the $j$th-$d$ of $\vec{SM}(sp_i)$, while $C$ represents for the number of classes.

Further, based on CLR unit, each superpixel is inferred to obtain class label $Cls(sp_i)$ and corresponding score $Sco(sp_i)$. In addition, superpixels whose score are less than a threshold (set as $0.5$ in experiments) are marked and ignored when network training. Noted that since each class performs semantic learning, CLR are operated directly for all categories in iteration process. With the generated annotations, a new round of learning can be operated. It can be analyzed that the segmentation network is used to learn common class semantics from different images, while CLR provides optimization guidance based on prior within each single image. By alternately iterating the two operation, inter- and intra-image information can be fully integrated, thus better performance can be achieved.

\section{Experiments}\label{sec: exp}
The proposed method is evaluated on Cityscapes \cite{cordts2016cityscapes}, a dataset which contains $5000$ images with fine annotations. Here, $2975$ images are divided to \textit{train} set, while $500$ and $1525$ samples are classified to \textit{val} and \textit{test} set. In official regulations, $19$ classes belonging to $7$ categories should be evaluated. For task setting, $19$ pixels belongs to different classes from $5$ images are labeled. To achieve optimal feature settings, $12$ classes are divided as objects: fence, pole, traffic light, traffic sign, as well as all types in human and vehicle categories, while the remaining $7$ classes are defined as scenes. The intersection-over-union averaged on both classes and categories (named as mIoU Class and mIoU Cate.) are used to evaluate performance of each method.

\begin{table}[t]\footnotesize 
	\setlength{\tabcolsep}{2.5 pt}
	\caption{Comparison of methods under different weakly supervised conditions on Cityscapes \textit{val} and \textit{test} set.}
	\vspace{-2.1mm}
	\begin{center}
		\begin{tabular}{l | c | c c | c c}
			\hline
			\cline{1-6} 
			\multirow{2}*{Methods} & \multirow{2}*{Annotations} &
			\multicolumn{2}{c|}{Results on \textit{val} set} & \multicolumn{2}{c}{Results on \textit{test} set} \\
			& & mIoU Class & mIoU Cate. & mIoU Class & mIoU Cate. \\ \hline
			SEC \cite{kolesnikov2016seed} & Image Tags & 2.3 & 7.1 & 2.4 & 7.2\\
			CCNN \cite{pathak2015constrained} & Image Tags & 7.3 & 16.3 & 7.2 & 17.2\\
			PSA \cite{ahn2018learning} & Image Tags & 21.6 & 39.0 & 21.2 & 40.2\\
			\hline
			Ours & One Pixel per Class & \textbf{23.8} & \textbf{46.1} & \textbf{26.6} & \textbf{48.3}\\
			\cline{1-6}
		\end{tabular}
	\end{center}
	\label{tab:val_test}
	\vspace{-8mm}
\end{table}

\subsection{Performance of the proposed framework}
\subsubsection{Comparison with State-of-the-art Methods:}
To the best of our knowledge, the proposed task has not been discussed or illustrated experimental results under similar settings. Meanwhile, due to the difference of labeling conditions, it is difficult to introduce other methods based on same initial annotations for fair comparison. Therefore, we compared our work with other methods which are based on fully image-level annotations. Here, three state-of-the-art weakly supervised methods are reproduced based on Cityscapes: SEC \cite{kolesnikov2016seed}, CCNN \cite{pathak2015constrained} and PSA \cite{ahn2018learning}. Table~\ref{tab:val_test} shows each method's performance on \textit{val} and \textit{test} set of Cityscapes respectively. By comparison, it can be seen that our method achieves better results under more lightweight condition, which proves both rationality of task setting and effectiveness of the proposed learning framework. Figure~\ref{fig:vis_result} illustrates several visualized results of the generated pseudo annotations and inferred semantic segmentation maps, which show that high quality results can be obtained based on the proposed learning framework.

\subsubsection{Analysis of Failure Cases:}
It can be seen that the method has better performance for both large scenes (such as flat, nature) or small objects (vehicles). However, the results could achieve further improved on several categories that are unstable to both low-level and high-level features. For example, as the category of human in natural scenes does not appear in ImageNet dataset, the performance of optimal feature encoding based on semantic feature is relatively low. While for fairness of evaluation, we do not introduce additional semantic prior information which could further improve the performance.

\subsection{Ablation Studies}
\subsubsection{Annotation with Optimal Feature Setting:} 

\begin{table}[t]\footnotesize 
	\setlength{\tabcolsep}{5.6 pt}
	\caption{The effectiveness of Optimal Feature Setting on Cityscapes \textit{val} set.}
	\vspace{-5.1mm}
	\begin{center}
		\begin{tabular}{l | c c c c}
			\hline
			\cline{1-5}
			Methods & Precision & Recall & mIoU Class &  mIoU Cate. \\ 
			\hline
			Only Imaging & 19.0 & 17.3 & 10.2 & 26.9 \\
			Only Semantics & 14.2 & 30.8 & 8.7 & 18.3 \\
			Single Enc. & 33.0 & 25.3 & 15.6 & 38.9 \\
			Ours & \textbf{33.8} & \textbf{31.5} & \textbf{20.4} & \textbf{41.5} \\
			\cline{1-5}	
		\end{tabular}%
	\end{center}
	\label{tab:seedcomp}
	\vspace{-8mm}
\end{table}

To evaluate the rationality of optimal feature setting for objects and scenes representation, we compared learning performance on Cityscapes \textit{val} set based on different pseudo annotations. First, based on the same semantic and imaging feature representation setting method, we constructed two types of expressions for all $19$ classes in Cityscapes. Then, to take part in the process of generate pseudo annotations, semantic feature (named as Only Semantics) and imaging feature (named as Only Imaging) based representation for classes are selected respectively. In addition, to evaluate the performance of multi overlapping slices fusion (MOSF), semantic feature directly encoded from entire image based on CAM (named as Single Enc.) is also compared. Note that in addition to the difference in feature representation, parameter settings and strategy utilized in pseudo labeling and semantic learning are consistent. Based on the annotations generated via different feature expression methods, semantic segmentation networks can be trained respectively and provide inference results. Table~\ref{tab:seedcomp} shows the performance which obtained with once CNN-based learning process. It can be seen that experiments based on optimal feature setting achieves better pseudo annotations and semantic segmentation results.

\subsubsection{Optimization via CLR Unit:} 

We further examine the optimization performance via context and location based refinement unit. Here, average precision and recall are introduced to evaluate the quality of pseudo labels. Table~\ref{tab:ilmcomp} provides the performance on Cityscapes \textit{train} set when different process of refinement are utilized individually or in combination, while annotations generated without CLR are also listed as baseline. The results prove that both precision and recall of pseudo annotations could be improved to a significant extent via CLR unit. Moreover, it should be note that only naive rule and rough threshold are introduced as context and location prior in CLR, it does not perform any statistical calculations or special process for Cityscapes, so the unit can be universally applied to other tasks under driving scenes.

\begin{table}[t]\footnotesize 
	\setlength{\tabcolsep}{6 pt}
	\caption{Performance of context-location based refinement on Cityscapes \textit{train} set.}
	\vspace{-5.1mm}
	\begin{center}
		\begin{tabular}{c c | c c}
			\hline
			\cline{1-4}
			Context based Mod. & Location based Mod. & Precision & Recall \\ 
			\hline
			& & 26.0 & 17.0 \\
			\checkmark & & 28.8 & 16.8 \\
			& \checkmark & 31.7 & 19.7 \\
			\checkmark & \checkmark & \textbf{34.4} & \textbf{19.9} \\
			\cline{1-4}	
		\end{tabular}%
	\end{center}
	\label{tab:ilmcomp}
	\vspace{-5mm}
\end{table}

\subsubsection{Alternative Iteration Learning Framework:} 

In order to verify the effectiveness of alternate iteration structure, which combines the process of CNN based common semantic learning and context-location based refinement, we evaluated the performance of trained model from each iteration on \textit{val} set. Here, both mIoU Class and mIoU Cate. are listed. For comparison, we also test iteration method which directly utilizes inference results from period CNN-based learning stage without CLR as pseudo annotations for new rounds of training. Note that the two iteration process start with same initial seeds. Table~\ref{tab:iter} shows the comparison results. We observe that even if CNN-based learning stage may introduce several noise labels, CLR unit can effectively remove wrong annotations and provide refined results. So it can be concluded that the framework can better improve the performance via stepping process.

\begin{table}[t]\footnotesize 
	\setlength{\tabcolsep}{4.7 pt}
	\caption{Results of the iteration process. Each iteration are evaluated on Cityscapes \textit{val} set with VGG-16 network.}
	\vspace{-2.1mm}
	\begin{center}
		\begin{tabular}{l|c c|c c}
			\hline
			\cline{1-5} 
			\multirow{2}*{Methods} &
			\multicolumn{2}{c|}{Iteration w/o CLR } & \multicolumn{2}{c}{Iteration w/ CLR} \\
			& mIoU Class & mIoU Cate. & mIoU Class & mIoU Cate. \\ \hline
			Initial Seed & - & - & 20.4 & 41.5 \\
			Iteration 1 & 20.8 & 42.7 & 22.3 & 43.9  \\
			Iteration 2 & 21.2 & 43.3 &  23.1 & 45.2 \\
			Iteration 3 & 21.0 & 43.0 & 23.8 &  46.1 \\
			\cline{1-5}
		\end{tabular}
	\end{center}
	\label{tab:iter}
	\vspace{-8mm}
\end{table}

\section{Conclusion} \label{sec: cons}
In this paper, a new weakly supervised condition is proposed for semantic segmentation task under complex driving scenes, in which only one annotated pixel is provided for each class. As visual appearance can not be better obtained based on image tags under complex scenes which include a large number of categories, the proposed setting provides a reasonable and lightweight annotation condition. Under this circumstance, three-step pseudo labeling method is presented, which gradually achieves optimal feature representation for each category, image inference and context-location based refinement. Here, semantic and imaging prior of driving scene sis fully utilized to establish discriminative expression for each types of object and scene, as well as to implement modification based on intra-image restriction. Further, alternative iteration learning structure is established with segmentation network and CLR unit. By integrating processes of inter-image common class semantic learning and intra-image context-location based refinement, the performance can be gradually improved. Experiments on Cityscapes prove both the rationality of task settings and effectiveness of the proposed learning framework.

\bibliographystyle{splncs04}
\bibliography{egbib_arXiv}


\end{document}